\documentclass[11pt]{article}

\usepackage[final]{acl}

\makeatletter

\renewenvironment{figure*}{\@dblfloat{figure}}{\end@dblfloat\if@firstcolumn\linenumbers\fi}
\makeatother

\usepackage{times}
\usepackage{latexsym}

\usepackage[T1]{fontenc}

\usepackage[utf8]{inputenc}

\usepackage{microtype}

\usepackage{graphicx}

\usepackage{amsmath,amssymb,amsfonts}
\usepackage{algorithmic}
\usepackage{algorithm}
\usepackage{booktabs}
\usepackage{xcolor}
\usepackage{tikz}
\usetikzlibrary{arrows.meta,positioning,shapes.geometric,decorations.pathmorphing,calc}

\title{Summarize, Judge, Refine:\\Decoupled Content Understanding and Policy Learning\\for Multimodal Content Moderation}

\author{
Zeeshan Ahmed, Yang Qin and Hanqing Huang \\
Meta AI \\
{\small\texttt{\{ahzee, yangqin, hanqinghuangy\}@meta.com}}
}

\begin{document}
\maketitle

\begin{abstract}
Content moderation systems traditionally entangle multimodal understanding with policy-specific classification, requiring full pipeline retraining for every policy change and suffering from label scarcity since multimedia cannot be meaningfully augmented. We propose \textbf{Summarize-Judge-Refine (SJR)}, a two-model architecture that decouples these concerns via a natural language interface: a multimodal Content Model produces structured text summaries, and a text-only Policy Model classifies them against policy definitions. An iterative co-training loop refines the Content Model via GRPO to produce policy-relevant summaries, while text-space augmentation generates adversarial summary variants---an augmentation pathway impossible on raw multimedia---enabling few-shot policy bootstrap. Every decision is grounded in a human-readable summary, providing interpretability as a structural byproduct. On misleading advertisement detection, SJR achieves +23.6\% relative non-misleading F1 over a zero-shot chain-of-thought baseline, outperforming end-to-end SFT, STaR/RFT, and RLFT. Notably, a variant trained on \emph{zero real violating examples}---with all positive-class data synthetically generated---matches the full-data model within 0.2\% relative on violating F1, demonstrating that new policies can launch without any real violation data.
\end{abstract}

\section{Introduction}
\label{sec:intro}

Online platforms host billions of pieces of multimedia content---videos, images, and text---that must be evaluated against hundreds of evolving content policies. The scale and diversity of this challenge demands automated content moderation systems that are accurate, interpretable, and rapidly adaptable to new policies. For example, an advertisement might show a trustworthy product page, but its landing page delivers a deceptive phishing site; detecting this requires understanding both modalities and reasoning about their consistency against a ``misleading advertising'' policy.

The dominant paradigm trains end-to-end multimodal classifiers that map raw content directly to violation decisions \citep{kiela2020hateful, chatzakou2019detecting}. While effective in high-resource settings, this approach suffers from two fundamental limitations:

\begin{enumerate}
    \item \textbf{Tight coupling.} Content understanding and policy logic are entangled in model weights. Every policy change---from definition updates to entirely new categories---requires retraining the full multimodal pipeline, a process that typically takes weeks of data collection, labeling, and training. This creates an operational bottleneck: platforms maintain hundreds of policies that evolve continuously.

    \item \textbf{Cold-start problem.} New policies launch with scarce labeled data. Multimedia data is \emph{fundamentally resistant to augmentation}: rotating an image or adding Gaussian noise does not alter its semantic content, cropping risks removing the violating region, and generating realistic violating videos raises both technical and ethical challenges. This asymmetry---text is easy to augment, multimedia is not---is the central observation that motivates our approach.
\end{enumerate}

\paragraph{Our insight.} Natural language can serve as a \emph{complete information bridge} between content understanding and policy classification. Projecting multimodal content into structured text summaries yields modularity (independent training), augmentability (NLP toolkit for policy learning), interpretability (summaries as rationales), multilingual coverage (any-language content normalized to a common language), and training efficiency (policy iteration requires SFT on a text-only model rather than the full multimodal pipeline).

\paragraph{Key challenge.} A generic summarizer produces \emph{policy-agnostic} descriptions that omit the details needed for correct classification. We address this through \emph{iterative co-training}: a reinforcement learning loop refines the Content Model so that its summaries progressively emphasize policy-relevant signals, while the Policy Model learns from increasingly informative summaries---a cooperative dynamic where both models teach each other through the text bridge.

\paragraph{Contributions.} Our contributions are as follows:
\begin{enumerate}
    \item A \textbf{two-model architecture} (\textbf{Content Model} $\rightarrow$ \textbf{Text Summary} $\rightarrow$ \textbf{Policy Model}) that formally decouples multimodal content understanding from policy classification via a natural language interface, enabling independent training and modular policy updates, with interpretability as a structural byproduct (\S\ref{sec:method}).
    \item An \textbf{iterative co-training algorithm} in which \emph{both} the Content Model and Policy Model evolve across rounds: the trained Policy Model serves as an autorater whose classification accuracy provides the reward signal for GRPO-based \citep{shao2024deepseekmath} refinement of the Content Model, which in turn produces improved summaries that the Policy Model retrains on---a cooperative dynamic structurally distinct from single-model self-training (STaR/RFT), fixed-reward RL, or frozen-teacher distillation (\S\ref{sec:method}).
    \item A \textbf{text-space augmentation framework} that generates adversarial and counterfactual summary variants from an augmentation model, enabling few-shot policy bootstrap from as little as non-violating examples only---an augmentation pathway that is impossible on raw multimedia. We show that a Policy Model trained on \emph{zero real violating examples} matches a full-data model within 0.2\% relative on violating F1 (\S\ref{sec:augmentation}).
    \item Comprehensive experiments on misleading advertisement detection, comparing SJR against zero-shot prompting, chain-of-thought reasoning \citep{wei2022chain}, end-to-end SFT classifiers, self-training (STaR/RFT; \citealp{zelikman2022star, yuan2023rft}), and reinforcement learning from task feedback (RLFT), demonstrating that SJR with text-space augmentation achieves superior accuracy while providing interpretable decisions (\S\ref{sec:results}).
\end{enumerate}

\section{Related Work}
\label{sec:related}

\subsection{End-to-End Content Moderation}

Traditional content moderation trains classifiers directly on multimodal inputs with policy-specific labels. Early work used CNNs for image-based detection \citep{chatzakou2019detecting}, while recent approaches leverage multimodal transformers \citep{kiela2020hateful} and causal debiasing methods for multimodal clickbait detection \citep{zheng2024clickbait}. Multi-task approaches share a backbone across policies but still couple the learned representation to the joint policy distribution \citep{schuster2021get}. End-to-end supervised fine-tuning (SFT) of large multimodal models has become a strong baseline: the entire model is trained to map raw inputs directly to policy labels, achieving high accuracy when sufficient labeled data is available. However, every new policy or label-distribution shift requires retraining the full pipeline.

More recently, LLMs have been deployed for content moderation at industrial scale. \citet{qiao2024scaling} describes a system for scaling LLM-based reviews for advertising policy violations in online advertising---a setting closely related to ours. \citet{oak2025reranking} use LLMs for re-ranking social media content to reduce exposure to harmful material at platform scale. These systems achieve strong performance but remain end-to-end: content perception and policy logic are entangled in the same model weights, they leave little to no room for multimedia data augmentation, and extending to new policies requires retraining. SJR explicitly separates content summarization from policy classification and optimizes both iteratively.

\subsection{Policy-as-Prompt and LLM-Based Moderation}

Recent work explores encoding policies as natural language prompts to LLMs. \citet{palla2025policy} formalizes the ``Policy-as-Prompt'' framework, showing LLMs can moderate content given policy descriptions as context. \citet{bonagiri2025safer} demonstrates scalable few-shot moderation using LLMs. Chain-of-thought (CoT) prompting \citep{wei2022chain} further improves LLM-based classification by eliciting intermediate reasoning steps before a final judgment. Relatedly, NLI-based approaches frame violation detection as textual entailment, where the content serves as a premise and the policy description as a hypothesis \citep{yin2019benchmarking, calabrese2022nli}, naturally supporting zero-shot transfer to new policies. These approaches achieve implicit decoupling through the prompt or entailment interface.

\subsection{Self-Training, Iterative Refinement, and RL Fine-Tuning}

Self-training methods iteratively improve a model by training on its own high-confidence predictions. STaR (Self-Taught Reasoner; \citealp{zelikman2022star}) bootstraps chain-of-thought reasoning by filtering for correct-answer traces and retraining, progressively improving reasoning quality. RFT (Rejection Fine-Tuning; \citealp{yuan2023rft}) generalizes this idea by sampling multiple candidate solutions and retraining on correct ones. RAFT (Reward rAnked FineTuning; \citealp{dong2023raft}) bridges SFT and RL by ranking model-generated samples with a reward model and fine-tuning on top-ranked outputs. Reinforcement learning from human feedback (RLHF; \citealp{ouyang2022training}) and its AI-feedback variant RLAIF \citep{bai2022constitutional, lee2023rlaif} train language models using preference-based reward signals, typically with a frozen reward model. Reinforcement learning from task feedback (RLFT) directly optimizes model outputs against task-specific reward functions (e.g., string-match accuracy), avoiding the need for preference data altogether.

SJR's co-training loop relates to these paradigms but differs in a structural way: \emph{both} the generator (Content Model) and the evaluator (Policy Model) are iteratively retrained. In RLHF/RLAIF, the reward model is fixed; in STaR/RFT, there is a single model that self-improves. In SJR, the Policy Model provides a reward signal to the Content Model, is then retrained on the Content Model's improved summaries, and produces an updated reward signal---creating a \emph{cooperative co-training} dynamic. Unlike knowledge distillation \citep{hinton2015distilling}, where the teacher is frozen, or GANs \citep{goodfellow2014generative}, where the generator and discriminator compete adversarially, SJR's two models cooperate: the Policy Model rewards the Content Model for useful summaries while itself learning from them. We adopt GRPO \citep{shao2024deepseekmath}, which eliminates the need for a separate critic network by computing advantages relative to a group of sampled outputs.

\subsection{Data Augmentation for Low-Resource Classification}

Text augmentation techniques---synonym replacement and simple edits \citep{wei2019eda}, back-translation \citep{sennrich2016improving}, LLM-based synthetic generation \citep{schick2021generating}, and counterfactual augmentation \citep{kaushik2020learning}---are effective for low-resource NLP tasks. \citet{zeng2024synthetic} recently showed that multimodal LLMs can generate synthetic training data for misinformation detection, validating the broader idea of LLM-generated augmentation for content moderation. However, these techniques operate on text and cannot be directly applied to raw multimedia: rotating a video frame does not change its semantic content, and generating realistic violating images raises technical and ethical challenges.

\textbf{SJR's text-space augmentation} resolves this asymmetry by introducing an Augmentation Model that takes raw multimodal content and a policy description as input, and generates adversarial summary variants with specific misleading elements injected---producing training pairs whose \emph{outputs} are text summaries even though their \emph{inputs} are real multimedia. This mechanism helps enable few-shot policy bootstrap: multimodal examples---even non-violating ones only---are fed to the Augmentation Model, which produces thousands of diverse text-space examples for the Policy Model.

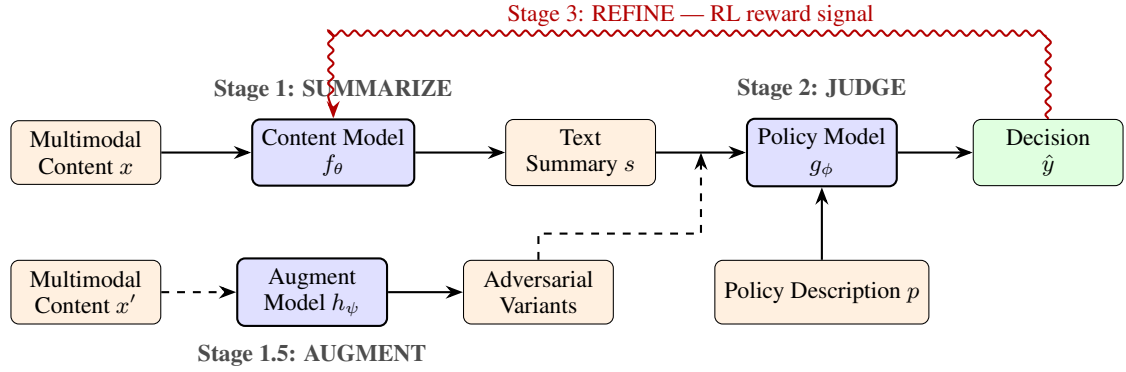
\begin{figure*}[t]
\centering
\resizebox{0.92\textwidth}{!}{%
\begin{tikzpicture}[
    >=Stealth,
    box/.style={draw, rounded corners=3pt, minimum height=0.85cm, minimum width=2.0cm, align=center, font=\small},
    model/.style={box, fill=blue!12, thick},
    data/.style={box, fill=orange!12},
    output/.style={box, fill=green!12},
    stage/.style={font=\footnotesize\bfseries, text=gray!60!black},
    arr/.style={->, thick},
    darr/.style={->, thick, dashed},
]

\node[data] (input) {Multimodal\\Content $x$};
\node[model, right=1.2cm of input] (content) {Content Model\\$f_\theta$};
\node[data, right=1.2cm of content] (summary) {Text\\Summary $s$};

\draw[arr] (input) -- (content);
\draw[arr] (content) -- (summary);

\node[stage, above=0.1cm of content] {Stage 1: SUMMARIZE};

\node[data, below=1.0cm of input] (auginput) {Multimodal\\Content $x'$};
\node[model, right=1.0cm of auginput] (augment) {Augment\\Model $h_\psi$};
\node[data, right=1.0cm of augment] (augsummary) {Adversarial\\Variants};

\draw[darr] (auginput) -- (augment);
\draw[arr] (augment) -- (augsummary);

\node[stage, below=0.1cm of augment] {Stage 1.5: AUGMENT};

\node[model, right=1.2cm of summary] (policy) {Policy Model\\$g_\phi$};
\node[output, right=1.0cm of policy] (decision) {Decision\\$\hat{y}$};

\node[data] (policydesc) at (policy |- augment) {Policy Description $p$};

\coordinate (midpoint) at ($(summary.east)!0.5!(policy.west)$);
\draw[arr] (summary) -- (policy);
\draw[darr] (augsummary.north) -- ++(0,0.35) -| (midpoint);
\draw[arr] (policy) -- (decision);
\draw[arr] (policydesc) -- (policy);

\node[stage, above=0.1cm of policy] {Stage 2: JUDGE};

\draw[arr, color=red!70!black, thick,
    decorate, decoration={snake, amplitude=1.0pt, segment length=5pt, post length=3pt}]
    (decision.north) -- ++(0,1.1)
    -| node[pos=0.25, above, font=\footnotesize, text=red!70!black]
       {Stage 3: REFINE --- RL reward signal}
    (content.north);

\end{tikzpicture}%
}
\caption{The Summarize-Judge-Refine (SJR) architecture. \textbf{Stage~1}: a multimodal Content Model projects raw multimedia into a structured text summary. \textbf{Stage~1.5} (optional): an Augment Model generates adversarial and counterfactual summary variants. \textbf{Stage~2}: a text-only Policy Model classifies summaries against a policy description. \textbf{Stage~3}: the Policy Model's accuracy serves as a reward to refine the Content Model via GRPO, with a KL penalty anchoring it to pretrained behavior. Dashed arrows indicate optional augmentation. Stages repeat iteratively.}
\label{fig:architecture}
\end{figure*}

\section{Method: Summarize-Judge-Refine (SJR)}
\label{sec:method}

\subsection{Architecture}

Given multimodal content $x \in \mathcal{X}$, a policy description $p$, and a violation label $y \in \{0,1\}$, the standard approach learns a monolithic function $h: \mathcal{X} \times \mathcal{P} \rightarrow \{0,1\}$ that entangles content perception with policy reasoning. SJR instead decomposes this into two models connected exclusively through natural language text (Figure~\ref{fig:architecture}).

\paragraph{Content Model $f_\theta$.} A multimodal language model that maps raw multimedia content to a tuple of structured textual summaries: $s = (s_1, s_2, \ldots, s_n) = f_\theta(x)$, where each component is a natural language string describing one facet of the content (e.g., the on-platform and off-platform sides). Initialized from a pretrained multimodal LLM, $f_\theta$ ingests raw visual, auditory, and textual signals and produces summaries in sufficient detail for downstream policy classification. In the initial round, these summaries are generic---a general-purpose captioning of what the model observes. Through iterative co-training (\S\ref{sec:cotraining}), the summaries progressively emphasize policy-relevant signals.

\paragraph{Policy Model $g_\phi$.} A text-only classifier that takes a summary $s$ and policy description $p$ as input, outputting a binary verdict:
\begin{equation}
    \hat{y} = g_\phi(s, p)
\end{equation}
Designed to be a lightweight text-only LLM (see \S\ref{sec:experiments}), $g_\phi$ operates entirely in text space. Its input is a pair of structured summaries (e.g., an on-platform content summary and an off-platform landing-page summary) together with a policy definition prompt.

\paragraph{Natural Language Interface.} The summary $s$ is the \emph{only} information channel between the two models. The Policy Model never observes raw pixels, audio, or video frames. This strict separation enforces five properties: (i) \textbf{modularity}---the Content Model and Policy Model can be updated independently; (ii) \textbf{augmentability}---the Policy Model's training data (text summaries) can be augmented even when the underlying multimedia cannot; (iii) \textbf{interpretability}---the summary constitutes a human-readable rationale for every decision; (iv) \textbf{multilinguality}---the Content Model absorbs language diversity by summarizing any-language content into a common language, so a single Policy Model handles all languages without per-language labeled data; and (v) \textbf{training efficiency}---policy iteration requires fine-tuning only a text-only model instead of the full multimodal pipeline.

\subsection{Training Algorithm: Iterative Co-Training}
\label{sec:cotraining}

We train SJR through an iterative loop of three stages plus an optional augmentation step (see Algorithm~\ref{alg:sjr} in Appendix~\ref{sec:algorithm} for full pseudocode). Each round $k$ generates summaries with the current Content Model, trains the Policy Model on those summaries (plus augmented data), and then refines the Content Model via GRPO \citep{shao2024deepseekmath} using the Policy Model's accuracy as reward. For each input, $G$ candidate summaries are sampled, scored by the Policy Model, and group-relative advantages are computed. The Content Model is then updated via a clipped policy gradient with a KL penalty anchoring it to the pretrained distribution. The cooperative dynamic is important: the Content Model learns to describe content in a way that makes the Policy Model's job easier, and the Policy Model learns to classify from increasingly informative summaries.

\paragraph{Computational design.} Each stage involves only light fine-tuning (one or two epochs), not full retraining. In practice, 2--3 rounds are needed for the co-training loop to stabilize, as the summaries quickly become policy-relevant and subsequent rounds yield diminishing returns.

\subsection{Reward Design}

The reward for each generated summary is based on whether the trained Policy Model, acting as an autorater, correctly classifies the summary:
\begin{equation}
    R(s, p, y) = \mathbb{1}\left[\hat{y} = y\right]
\end{equation}
where $\hat{y} = g_\phi(s, p)$ is the Policy Model's predicted label (extracted via string matching from its generated output) and $y$ is the gold label. The reward is binary: 1 if the prediction matches the label, 0 otherwise. This rule-based reward uses the same string-match mechanism as standalone RLFT, but with a structural difference: the reward signal flows through the \emph{trained Policy Model} operating on text summaries, not through the Content Model itself. The Policy Model thus serves as an autorater---its classification accuracy on the generated summary determines whether that summary is useful for downstream decision-making.


\subsection{Text-Space Augmentation for Few-Shot Policy Bootstrap}
\label{sec:augmentation}

The natural language interface enables a powerful data augmentation strategy that is \emph{structurally impossible} on raw multimedia. For a new policy $p_{\text{new}}$ with limited---or even exclusively non-violating---multimodal examples, we generate diverse text-space training data for the Policy Model using an Augmentation Model that takes raw multimodal content as input.

\paragraph{Adversarial summary generation.} The primary augmentation mechanism uses an \textbf{Augmentation Model} $h_\psi$ (a capable multimodal LLM) to generate adversarial summary variants. Given the raw multimedia content $x$ and the policy description $p$, $h_\psi$ produces multiple counterfactual summaries that represent plausible variations of the content:

\begin{enumerate}
    \item \textbf{Misleading variants}: $h_\psi$ modifies the summary to introduce specific deceptive elements---contradictory claims between the on-platform and off-platform descriptions, exaggerated promises, or misdirected calls-to-action---producing labeled positive (violating) examples anchored on real content patterns.

    \item \textbf{Benign variants}: $h_\psi$ generates paraphrases that preserve the non-violating semantics while varying surface form, producing diverse negative examples.

    \item \textbf{Borderline variants}: $h_\psi$ generates edge cases with subtle discrepancies, forcing the Policy Model to learn fine-grained decision boundaries instead of superficial cues.
\end{enumerate}

Each augmented example carries a label assigned by the Augmentation Model based on whether the generated variant constitutes a violation under the policy definition. This produces a $4$--$6\times$ expansion per input example, yielding $5$--$7n$ total training examples from $n$ multimodal seeds (which need not include any violating examples)---all without additional data collection or labeling.

\paragraph{Zero-positive policy bootstrap.} A particularly powerful consequence of text-space augmentation is that new policies can launch with \emph{zero real positive examples}. The Augmentation Model takes real non-violating content and generates synthetic violating variants, producing a training set where all negative examples are real and all positive examples are synthetically generated. This eliminates the positive-label collection bottleneck: a new policy can ship as soon as a small set of verified legitimate examples is available, with the Policy Model bootstrapping its understanding of violations entirely from synthetic contrast. We evaluate this extreme few-shot scenario in \S\ref{sec:results}.

\paragraph{Why this is impossible on multimedia.} Generating a ``misleading variant'' of a real video advertisement---modifying the landing page to contain contradictory claims while keeping visual fidelity---is technically intractable. Text summaries, by contrast, can be freely modified, paraphrased, and counterfactually edited while maintaining coherence. The natural language interface transforms an impossible multimedia augmentation problem into a straightforward text generation task.

\section{Experimental Setup}
\label{sec:experiments}

\subsection{Task: Misleading Advertisement Detection}

We evaluate SJR on \textbf{misleading advertisement detection}, a multimodal content moderation task that requires comparing what an advertisement promises (\emph{on-platform content}) against what its destination actually delivers (\emph{off-platform landing page}). An advertisement is \emph{misleading} (positive) if the landing page contradicts, misrepresents, or fails to deliver on the claims made in the ad creative; it is \emph{non-misleading} (negative) if the landing page is consistent with the advertisement. The exact decision boundary varies across applicable policies, making this a setting where SJR's modular policy updates are particularly valuable.

This task is a natural fit for SJR because it inherently requires \emph{cross-modal comparison}: the on-platform ad may contain images, video, and overlay text, while the off-platform landing page is a separate web document with its own text and visual elements. An end-to-end classifier must jointly encode both modalities and learn the comparison logic; SJR instead summarizes each side into text and delegates the comparison to the Policy Model.

\subsection{Dataset}

We use a dataset of paid advertisements collected from a major online advertising platform. Each example consists of:

\begin{itemize}
    \item \textbf{On-platform signals}: advertisement text and associated images/videos, OCR-extracted text from visual creatives, and the destination URL linking to the off-platform landing page.
    \item \textbf{Off-platform signals}: scraped landing-page text content and landing-page screenshots.
    \item \textbf{Label}: binary decision (misleading vs.\ non-misleading).
\end{itemize}

\paragraph{Training set.} Approximately 120K examples with knowledge-distilled labels (55\% non-misleading, 45\% misleading) obtained from a larger teacher model, supplemented with a small set of human-annotated examples to anchor label quality. Distilled labels are expected to contain inherent noise, which particularly affects RL-based methods.

\paragraph{Evaluation set.} A separate set of ${\sim}$2K examples with expert-corrected labels from domain specialists. Each example carries an importance weight reflecting estimated real-world prevalence, enabling both unweighted and weighted metric reporting. Multiple landing-page scrapes per advertisement support per-entity OR-aggregated evaluation (an advertisement is flagged if \emph{any} of its scrapes is classified as misleading).

\paragraph{Evaluation metrics.} We report precision, recall, and F1 for both the positive class (misleading) and the negative class (non-misleading). All metrics are weighted by estimated real-world prevalence and per-entity OR-aggregated. The primary metric is \textbf{weighted per-entity non-misleading F1}, which measures the system's ability to correctly identify legitimate advertisements---the operationally significant metric, since false positives would result in legitimate ads being incorrectly blocked.

\subsection{Baselines}
\label{sec:baselines}

We compare SJR against four families of baselines (Table~\ref{tab:baselines}), all evaluated on the same evaluation set for direct comparability.

\begin{table}[t]
\centering
\small
\begin{tabular}{@{}p{2.8cm}p{4.2cm}@{}}
\toprule
\textbf{Method} & \textbf{Description} \\
\midrule
\textbf{Zero-shot CoT} & Large multimodal LLM classifies directly from raw content with the policy encoded in the prompt, eliciting chain-of-thought reasoning \citep{wei2022chain} before the final verdict. No training. Serves as our prompt-only reference point. \\
\addlinespace
\textbf{End-to-end SFT} & Supervised fine-tuning of a multimodal LLM on the full training set, mapping raw content directly to a binary label. The standard approach. \\
\addlinespace
\textbf{STaR/RFT} & Self-Taught Reasoner / Rejection Fine-Tuning \citep{zelikman2022star, yuan2023rft}: iteratively generates CoT traces, filters for correct answers, and retrains on successful traces. Multiple rounds of self-bootstrapping. \\
\addlinespace
\textbf{RLFT} & Reinforcement learning from task feedback: optimizes a single model via RL with string-match accuracy as reward. Multiple rounds with fresh data samples. \\
\bottomrule
\end{tabular}
\caption{Baseline methods. All methods use the same base LLM family and are evaluated on the same evaluation set.}
\label{tab:baselines}
\end{table}

\paragraph{SJR variants.} We evaluate several configurations of SJR to isolate the contribution of each component:

\begin{itemize}
    \item \textbf{Two-step (no training)}: Content Model summarizes, Policy Model classifies---both using pretrained weights with no fine-tuning. Tests the architecture alone.
    \item \textbf{SJR (JUDGE only)}: Policy Model is fine-tuned on real summaries from the pretrained Content Model (no augmentation, no RL on the Content Model). Tests the value of training the Policy Model alone.
    \item \textbf{SJR + Augmentation}: Same as above, but the Policy Model is trained on real summaries \emph{plus} adversarial variants from the Augmentation Model. The Content Model remains pretrained (no RL).
    \item \textbf{SJR + Augmentation (multi-round)}: The full SJR loop with multiple co-training rounds: the Content Model is refined via GRPO (REFINE stage) using the Policy Model as autorater, and the Augmentation Model regenerates fresh adversarial summaries each round.
    \item \textbf{SJR + Augmentation (non-misleading only)}: Same as multi-round, but the seed data contains \emph{only non-misleading} examples---zero real violating examples. All positive-class data is synthetically generated via the zero-positive bootstrap mechanism (\S\ref{sec:augmentation}). Tests whether SJR can learn to detect violations without ever seeing a real one.
\end{itemize}

\subsection{Implementation Details}

All three models---Content, Augmentation, and Policy---use the same multimodal LLM capable of processing images, video, and text. The SJR architecture is designed to use a smaller, text-only model for the Policy Model, which would reduce inference cost and enable faster policy iteration; in our experiments, we use the same model for all three roles to isolate the architectural contribution from model-capacity effects. We leave the cost benefits of a smaller Policy Model to future work.

All fine-tuning---SFT, STaR/RFT, RLFT, and SJR---uses LoRA adapters (rank 8) for parameter-efficient training. For the \textbf{end-to-end SFT} baseline, the multimodal LLM is fine-tuned on the full training set, mapping raw content directly to a binary label. \textbf{STaR/RFT} runs 3--5 iterative rounds: in each round, the model generates multiple CoT traces per example at non-zero temperature, retains only traces that produce the correct final answer, and retrains on the filtered set. \textbf{RLFT} optimizes a single model via GRPO with binary string-match accuracy as reward over multiple rounds.

For \textbf{SJR}, GRPO uses a group size of $G=16$ candidate summaries per input. The Policy Model is trained for 3 epochs in the first round and 1 epoch in subsequent rounds. Content Model RLFT (when enabled) uses 1 epoch per round with a KL penalty coefficient $\lambda$ to prevent drift. All baselines and SJR variants use temperature $T=0$ (greedy decoding) at evaluation for reproducibility.

\section{Results}
\label{sec:results}

Table~\ref{tab:results} presents the main results. SJR with augmentation and multi-round co-training achieves the highest F1 on both classes: 79.25\% non-misleading F1 (+23.6\% relative over the zero-shot CoT baseline) and 93.73\% misleading F1 (+2.1\% relative), outperforming all baselines by a wide margin.

\begin{table*}[t]
\centering
\small
\begin{tabular}{@{}l ccc ccc@{}}
\toprule
& \multicolumn{3}{c}{\textbf{Non-Misleading}} & \multicolumn{3}{c}{\textbf{Misleading}} \\
\cmidrule(lr){2-4} \cmidrule(lr){5-7}
\textbf{Method} & P & R & F1 & P & R & F1 \\
\midrule
Zero-shot CoT (ref.) & 82.63 & 52.41 & 64.14 & 87.30 & 96.74 & 91.78 \\
End-to-end SFT & 76.80 & 73.94 & 75.34 & 92.38 & 93.39 & 92.88 \\
STaR/RFT & 80.66 & 72.28 & 76.24 & 90.15 & 91.96 & 91.04 \\
RLFT & 64.83 & \textbf{82.08} & 72.44 & 91.45 & 89.98 & 90.71 \\
\midrule
Two-step (no training) & \textbf{87.35} & 51.68 & 64.94 & 87.25 & \textbf{97.79} & 92.22 \\
SJR (JUDGE only) & 75.37 & 73.07 & 74.21 & 92.11 & 92.94 & 92.52 \\
SJR + Augmentation & 75.93 & 73.61 & 74.75 & 92.25 & 93.08 & 92.66 \\
SJR + Aug (multi-round) & 78.00 & 80.54 & \textbf{79.25} & \textbf{94.19} & 93.28 & \textbf{93.73} \\
\addlinespace
SJR + Aug (nm.\ only) & 78.40 & 78.03 & 78.21 & 93.51 & 93.64 & 93.58 \\
\bottomrule
\end{tabular}
\caption{Main results on misleading advertisement detection (\%). All metrics are weighted and per-entity OR-aggregated. Top group: baselines; bottom group: SJR variants, \S\ref{sec:experiments}. ``nm.\ only'' = trained with non-misleading examples only (zero real violating examples). ``multi-round'' = three rounds of co-training. Best value per column in \textbf{bold}.}
\label{tab:results}
\end{table*}

\paragraph{Baselines.} We report all relative improvements against the zero-shot CoT reference (non-misleading F1 64.14\%, misleading F1 91.78\%), a precision-heavy but recall-limited baseline (non-misleading recall 52.41\%). End-to-end SFT and STaR/RFT are the strongest single-model baselines, lifting non-misleading F1 by +17.5\% and +18.9\% relative, respectively. The two methods achieve these gains through different precision--recall trade-offs: STaR/RFT sharpens precision (80.66\%) at the cost of recall (72.28\%), while SFT improves both more evenly. On the misleading side, SFT lifts F1 by +1.2\% relative, while STaR/RFT and RLFT dip slightly below the reference ($-$0.8\% and $-$1.2\% relative). RLFT shows a more pronounced trade-off: it achieves the highest non-misleading recall of any method (82.08\%) but the lowest precision (64.83\%).

\paragraph{SJR component analysis.} The SJR rows isolate each component's contribution. Two-step inference without training (64.94\% non-misleading F1) is roughly on par with the zero-shot CoT baseline, confirming that the architectural decomposition is not inherently beneficial---the gains come from training. Training the Policy Model on real summaries (JUDGE only) raises non-misleading F1 to 74.21\%, approaching end-to-end SFT and demonstrating that a text-only classifier over summaries can nearly match a full multimodal model. Single-round augmentation adds a modest gain (74.75\%). The decisive improvement comes from multi-round co-training: the REFINE stage pushes non-misleading F1 to 79.25\%---a 4.5 percentage-point gain that accounts for roughly 30\% of SJR's total lift above the zero-shot CoT baseline.

\paragraph{Cold-start.} The non-misleading-only variant trains with \emph{zero real violating examples}---all positive-class data is synthetically generated via text-space augmentation (\S\ref{sec:augmentation}). After three co-training rounds, it achieves 78.21\% non-misleading F1 (+21.9\% relative over the zero-shot CoT baseline) and 93.58\% misleading F1 (+2.0\% relative), within 1.3\% and 0.2\% relative of the full-data variant. New policies can therefore launch as soon as a non-violating seed set is collected, removing the positive-label bottleneck entirely.

\section{Analysis}
\label{sec:analysis}

We analyze three aspects of SJR's behavior: the dynamics of multi-round co-training, the role of augmentation, and the convergence of the cold-start variant.

\begin{table}[t]
\centering
\small
\begin{tabular}{@{}c cc cc@{}}
\toprule
& \multicolumn{2}{c}{\textbf{Full data}} & \multicolumn{2}{c}{\textbf{NM only}} \\
\cmidrule(lr){2-3} \cmidrule(lr){4-5}
\textbf{Round} & NM-F1 & M-F1 & NM-F1 & M-F1 \\
\midrule
1 & 74.75 & 92.66 & 74.28 & 92.28 \\
2 & 72.25 & 92.25 & 76.31 & 93.85 \\
3 & \textbf{79.25} & \textbf{93.73} & \textbf{78.21} & \textbf{93.58} \\
\bottomrule
\end{tabular}
\caption{Per-round F1 (\%) for SJR + Augmentation with full training data vs.\ non-misleading-only (cold-start) data. NM = non-misleading, M = misleading. Best per column in \textbf{bold}.}
\label{tab:rounds}
\end{table}

\paragraph{Co-training dynamics.} Table~\ref{tab:rounds} shows a non-monotonic trajectory: non-misleading F1 dips from 74.75\% (R1) to 72.25\% (R2) before climbing to 79.25\% (R3). The R2 dip reflects a recall drop (73.61\% $\to$ 69.11\%) as the refined Content Model's summaries temporarily misalign with the Policy Model. By R3, the models co-adapt: recall recovers to 80.54\% while precision improves to 78.00\%, yielding a balanced profile unique among all methods. At least two REFINE iterations are needed for this balance to emerge.

\paragraph{Role of augmentation.} Single-round augmentation adds only 0.54 percentage points over JUDGE-only (74.75\% vs.\ 74.21\%), but it enables effective multi-round co-training by providing diverse adversarial variants that sharpen the Policy Model's boundary and yield a more informative reward signal for Content Model refinement. The cumulative contribution grows from +0.54 percentage points in Round~1 to +5.04 percentage points by Round~3.

\paragraph{Cold-start convergence.} The non-misleading-only variant converges toward full-data SJR across rounds, and unlike the full-data variant, its trajectory is monotonically increasing (no R2 dip). On misleading F1, it actually leads in R2 (93.85\% vs.\ 92.25\%), suggesting synthetic violations provide complementary diversity. By R3, both variants converge to within 0.15 percentage points on misleading F1.

\section{Conclusion}
\label{sec:conclusion}

We presented Summarize-Judge-Refine (SJR), a framework that decouples multimodal content understanding from policy learning through a natural language interface. On misleading advertisement detection, SJR outperforms all baselines on both the non-misleading and misleading F1 metrics, achieving +23.6\% and +2.1\% relative improvement over a zero-shot CoT baseline---gains that exceed end-to-end SFT, STaR/RFT, and RLFT by a substantial margin. The multi-round co-training loop is the decisive contributor, accounting for roughly 30\% of SJR's total lift.

SJR's most practically significant result is its cold-start capability: a variant trained with zero real violating examples achieves near-parity with the full-data system, removing the positive-label bottleneck that has historically delayed new policy launches. While SJR's two-model inference pipeline is more computationally expensive than a single end-to-end model, it can serve as a high-quality teacher for knowledge distillation into a smaller student model, combining SJR's accuracy and interpretability at training time with efficient single-model inference at serving time.

\section*{Limitations}

SJR introduces a two-model inference pipeline, increasing the number of LLM calls relative to a single end-to-end classifier. The architecture is designed to offset this through a smaller, text-only Policy Model; in our experiments, we use the same model for all three roles to isolate the architectural contribution, deferring the inference cost reduction to future work. The co-training loop exhibits non-monotonic dynamics---performance can dip in intermediate rounds before recovering---though in our experiments three rounds consistently yielded the best results.

The cold-start mechanism's effectiveness depends on the Augmentation Model's generation quality; we expect that more capable Augmentation Models will produce higher-quality synthetic violations, further improving cold-start performance.

Our evaluation covers a single content moderation task (misleading advertisement detection) on an internal dataset. While the architecture is domain-agnostic, generalization to other multimodal policy tasks remains to be demonstrated.

\bibliography{sjr_paper}

\clearpage
\appendix

\section{Training Algorithm}
\label{sec:algorithm}

\begin{algorithm}[h]
\caption{Summarize-Judge-Refine (SJR)}
\label{alg:sjr}
\begin{algorithmic}[1]
\REQUIRE Dataset $\mathcal{D} = \{(x_i, p_i, y_i)\}$, Content Model $f_\theta$, Policy Model $g_\phi$, Group size $G$, Rounds $K$, Clip range $\epsilon$, KL coefficient $\lambda$
\STATE Initialize $f_\theta \leftarrow$ pretrained multimodal LLM
\FOR{round $k = 1$ to $K$}
    \STATE \textbf{// Stage 1: SUMMARIZE}
    \FOR{each $(x_i, p_i, y_i) \in \mathcal{D}$}
        \STATE $s_i^{(k)} \leftarrow f_\theta(x_i)$ \COMMENT{Generate text summary}
    \ENDFOR
    \STATE \textbf{// Stage 1.5: AUGMENT (optional)}
    \STATE $\mathcal{D}_{\text{aug}}^{(k)} \leftarrow \text{AdversarialAugment}(\{x_i, p_i, y_i\})$
    \STATE \textbf{// Stage 2: JUDGE}
    \STATE Fine-tune $g_\phi$ on $\{(s_i^{(k)}, p_i, y_i)\} \cup \mathcal{D}_{\text{aug}}^{(k)}$
    \STATE Evaluate $g_\phi$ on validation set $\rightarrow$ metric$_k$
    \STATE \textbf{// Stage 3: REFINE (GRPO)}
    \FOR{each $(x_i, p_i, y_i) \in \mathcal{D}$}
        \STATE Sample $G$ summaries: $\{s_i^1, \ldots, s_i^G\} \sim f_\theta(\cdot | x_i)$
        \STATE Compute rewards: $r_i^g = R(s_i^g, p_i, y_i)$ for $g = 1, \ldots, G$
        \STATE Group-relative advantages: $\hat{a}_i^g = \frac{r_i^g - \mu(\{r_i^g\}_{g=1}^G)}{\sigma(\{r_i^g\}_{g=1}^G)}$
    \ENDFOR
    \STATE Update $f_\theta$ via clipped policy gradient:
    \STATE \quad $\mathcal{L} = -\mathbb{E}\left[\min\left(\rho \hat{a}, \text{clip}(\rho, 1{\pm}\epsilon)\hat{a}\right)\right] + \lambda \cdot \text{KL}(f_\theta \| f_{\theta_0})$
    \STATE \quad where $\rho = \pi_\theta(s|x) / \pi_{\theta_{\text{old}}}(s|x)$
    \IF{metric$_k$ converges}
        \STATE \textbf{break}
    \ENDIF
\ENDFOR
\RETURN $f_\theta, g_\phi$
\end{algorithmic}
\end{algorithm}

\section{Reward Function Extensions}
\label{sec:reward_extensions}

The binary reward $R$ used in our experiments (\S\ref{sec:method}) can be extended with additional terms for settings where faithfulness or fluency constraints are needed:
\begin{multline}
    R(s, p, y) = \alpha \cdot \mathbb{1}\left[\hat{y} = y\right] \\
    + \beta \cdot r_{\text{faithful}}(s, x) + \gamma \cdot r_{\text{fluency}}(s)
\end{multline}

\begin{itemize}
    \item \textbf{Faithfulness reward} $r_{\text{faithful}}$: measures whether the summary accurately reflects the actual content, preventing the Content Model from hallucinating policy-relevant details that do not exist in the input.

    \item \textbf{Fluency reward} $r_{\text{fluency}}$: ensures generated summaries remain coherent and human-readable, measured by perplexity under a reference language model.
\end{itemize}

In our experiments, we found that the binary accuracy reward combined with KL regularization was sufficient to maintain both faithfulness and fluency, so we set $\beta = \gamma = 0$. These extensions are included for generality and may be valuable in settings where the Content Model exhibits hallucination or degenerate outputs.

\section{Per-Round Precision, Recall, and F1}
\label{sec:per_round_details}

Table~\ref{tab:rounds_full} expands the per-round F1 results (Table~2 in the main text) with precision and recall for both classes, providing deeper insight into the co-training dynamics discussed in \S\ref{sec:analysis}.

\begin{table}[h]
\centering
\small
\resizebox{\columnwidth}{!}{%
\begin{tabular}{@{}ll ccc ccc@{}}
\toprule
& & \multicolumn{3}{c}{\textbf{Non-Misleading}} & \multicolumn{3}{c}{\textbf{Misleading}} \\
\cmidrule(lr){3-5} \cmidrule(lr){6-8}
\textbf{Variant} & \textbf{Rd} & P & R & F1 & P & R & F1 \\
\midrule
Full data & 1 & 75.93 & 73.61 & 74.75 & 92.25 & 93.08 & 92.66 \\
          & 2 & 75.68 & 69.11 & 72.25 & 91.09 & 93.43 & 92.25 \\
          & 3 & 78.00 & 80.54 & \textbf{79.25} & 94.19 & 93.28 & \textbf{93.73} \\
\midrule
NM only   & 1 & 73.43 & 75.14 & 74.28 & 92.60 & 91.96 & 92.28 \\
          & 2 & 85.55 & 68.87 & 76.31 & 91.30 & 96.56 & 93.85 \\
          & 3 & 78.40 & 78.03 & \textbf{78.21} & 93.51 & 93.64 & \textbf{93.58} \\
\bottomrule
\end{tabular}%
}
\caption{Full P/R/F1 (\%) per round for SJR + Augmentation. The Round~2 recall drop in the full-data variant (73.61\% $\to$ 69.11\%) and recovery in Round~3 (80.54\%) are visible. The NM-only variant shows a contrasting pattern: precision spikes in Round~2 (85.55\%) while recall drops (68.87\%), but both converge by Round~3.}
\label{tab:rounds_full}
\end{table}

\section{Prompt Templates}
\label{sec:prompts}

We provide the core instructions from each model's prompt template below, anonymized to remove platform-specific details.

\paragraph{Content Model (Stage 1: SUMMARIZE).} The Content Model receives both on-platform and off-platform signals and produces two independent summaries:

\begin{quote}
\small
\textit{You are an expert content analyst. You are shown two sides of a sponsored ad-and-landing-page pair: (A) the on-platform content the user sees before clicking, and (B) the off-platform landing page reached after clicking. Write TWO independent factual summaries---one of each side---AND classify each side's content category, so a downstream policy classifier can decide whether the landing page matches the on-platform promise. You must NOT make the misleading-link decision yourself.}

\textit{ON-PLATFORM SUMMARY (200--400 words): Write a cohesive factual preview of what the user is led to expect on the landing page after clicking.}

\textit{OFF-PLATFORM SUMMARY (200--400 words): Write a cohesive factual summary of what the page actually communicates.}
\end{quote}

\paragraph{Policy Model (Stage 2: JUDGE).} The Policy Model receives the summary pair and a policy description, outputting a binary verdict:

\begin{quote}
\small
\textit{You are an expert content moderator. Determine whether a URL is MISLEADING or NOT MISLEADING, given a JSON block describing both sides of the click: on\_platform\_summary and off\_platform\_summary. A link is MISLEADING only if the landing page is NOT semantically related to what the on-platform content promises. ``Semantically related'' means the same central topic, offer, or intent.}
\end{quote}

\paragraph{Augmentation Model (Stage 1.5: AUGMENT).} The Augmentation Model takes raw multimodal content and generates adversarial summary variants from a catalog of scenarios:

\begin{quote}
\small
\textit{You are an expert adversarial data generator. Given a real landing page and on-platform content, generate ONLY the adversarial off-platform summaries that are PLAUSIBLE for this specific input. Each summary should be either: (1) HARD NEGATIVE---not misleading, but easily confused as misleading (causes false positives), or (2) HARD POSITIVE---is misleading, but easily confused as not-misleading (causes false negatives). Pick only scenarios that genuinely fit this LP's brand, vertical, format, and content. Quality over quantity.}
\end{quote}

The adversarial scenario catalog spans hard-negative scenarios (cases that are not misleading but are easily confused as misleading, causing false positives) and hard-positive scenarios (genuinely misleading cases that are easily confused as not misleading, causing false negatives), covering offer-level mismatches, brand-confusion cases, and category- or redirect-based substitutions, among others.

\end{document}